\documentclass[runningheads]{llncs}
\usepackage[T1]{fontenc}
\usepackage{graphicx}
\usepackage{amsmath}
\usepackage{multirow}
\usepackage{booktabs}
\usepackage{float}
\usepackage{amsfonts}

\begin{document}
\title{Representation Learning of Auxiliary Concepts for Improved Student Modeling and Exercise Recommendation}
\author{Yahya Badran\inst{1,2}\orcidID{0009-0006-9098-5799} \and
Christine Preisach\inst{1,2}\orcidID{0009-0009-1385-0585}}
\authorrunning{Y. Badran \& C. Preisach}
\institute{Karlsruhe University of Applied Sciences, Moltekstr. 30, 76133 Karlsruhe, Germany \and
Karlsruhe University of Education, Bismarckstr 10,76133 Karlsruhe, Germany\\
\email{\{yahya.badran,christine.preisach\}@h-ka.de}}
\maketitle              % typeset the header of the contribution

\begin{abstract}
Personalized recommendation is a key feature of intelligent tutoring systems, typically relying on accurate models of student knowledge. Knowledge Tracing (KT) models enable this by estimating a student’s mastery based on their historical interactions. Many KT models rely on human-annotated knowledge concepts (KCs), which tag each exercise with one or more skills or concepts believed to be necessary for solving it. However, these KCs can be incomplete, error-prone, or overly general.

In this paper, we propose a deep learning model that learns sparse binary representations of exercises, where each bit indicates the presence or absence of a latent concept. We refer to these representations as auxiliary KCs. These representations capture conceptual structure beyond human-defined annotations and are compatible with both classical models (e.g., BKT) and modern deep learning KT architectures.

We demonstrate that incorporating auxiliary KCs improves both student modeling and adaptive exercise recommendation. For student modeling, we show that augmenting classical models like BKT with auxiliary KCs leads to improved predictive performance. For recommendation, we show that using auxiliary KCs enhances both reinforcement learning–based policies and a simple planning-based method (expectimax), resulting in measurable gains in student learning outcomes within a simulated student environment.

\keywords{Knowledge Tracing  \and Bayesian Knowledge Tracing \and Recommender System \and Reinforcement learning \and Representation Learning \and Deep Learning}
\end{abstract}
\section{Introduction}
Knowledge tracing (KT) is a fundamental task in educational data mining that involves modeling students’ mastery over time in order to predict their future performance. These models also serve as a core component in educational recommender algorithms \cite{dkvmn-ca,dkt,huo2020knowledge,sergey}. \footnote{This paper is an extended version of \cite{csedu}}

Many KT models rely on predefined knowledge concepts (KCs), which are discrete skills assumed to be required for solving a given exercise. Each exercise is typically tagged with one or more KCs based on expert annotations. However, predefined KCs may not fully capture the underlying complexities and latent structure of the learning process. They can be incomplete, noisy, or overly general, which can limit the accuracy of student models. For example, a human-defined KCs labeled "addition" may overlook the difference between "single-digit addition" and "addition involving decimals".

%Knowledge tracing (KT) is a fundamental task in educational data mining that involves modeling students’ mastery over time in order to predict their future performance. Accurate KT enables adaptive tutoring systems and informed instructional decisions. These models often rely on predefined KCs associated with each exercise. However, these predefined KCs may not fully capture the underlying complexities and latent structures of the learning process.

Recent advancements in representation learning have introduced the possibility of uncovering latent features through data-driven methods. This approach has been utilized in different areas of machine learning including education data mining~\cite{general_representation,pre_cont_embed2020}. By leveraging deep learning, we can identify hidden patterns and relationships that are not immediately apparent from predefined KCs alone. These representations can be further utilized in downstream tasks, such as improving the performance of simpler KT models like BKT, or enhancing adaptive recommendation strategies.

%, auxiliary KCs offer a more nuanced and comprehensive view of the latent knowledge space.

In this paper, we propose a Sparse Binary Representation Knowledge Tracing (SBRKT), a model that learns new tags analogous to human-defined KCs. These learned representations can be used by both traditional models, such as Bayesian Knowledge Tracing (BKT), and modern deep learning approaches. Specifically, we train a neural network to generate a sparse binary vector for each exercise. These vectors serve as the basis for deriving latent labels, which we refer to as auxiliary knowledge concepts (auxiliary KCs). In this binary representation, a value of one indicates the presence of an auxiliary KC, while a value of zero indicates its absence.

Although these auxiliary KCs do not carry explicit human labels, they can be integrated into downstream tasks to improve performance. In this work, we explore their use in two key tasks: (1) enhancing classical knowledge tracing models such as BKT, and (2) improving exercise recommendation algorithms.

Unlike pre-trained dense vector embeddings that are commonly used in deep learning, our learned representation can be integrated into non-deep learning models such as BKT by simply training it with the added auxiliary KCs. This approach caused BKT to outperform the original DKT on some benchmarks. With that, we help bridge the simplicity and interpretability of BKT with deep learning ability to capture complex dependencies.

%DKT represents a significant advancement in the field of knowledge tracing, offering a flexible and expressive framework for modeling student learning. Nonetheless, challenges related to interpretability and data requirements continue to underscore the value of classical models such as BKT. This work aims to bridge these approaches by transferring the representational strengths of deep learning into interpretable frameworks like BKT.

To utilize auxiliary KCs in recommendations, we note that a KT model can form a basis for a recommendation algorithm. In \cite{dkt}, they used a KT model to perform an expectimax algorithm where all the possible choices are tested on the model and the choice with the highest improvement is recommended. However, testing all exercises is impractical, instead the algorithm actually recommend KCs. Later, an exercise with the chosen KC can be recommended. In this paper, we adjusted the algortihm to recommend an auxiliary KC alongside the human-labeled KC which helps narrow down the set of possible exercises to recommend leading to a better performance.

Another approach to provide recommendations is the use of Reinforcement Learning (RL). In this approach a policy is optimized to recommend exercises. In this paper, we introduce a deep RL architecture that use a deep knowledge tracing (DKT)\cite{dkt} like model in the policy, which can incorporate KCs similar to the original DKT model. This helps incorporate our learned auxiliary KCs into the policy and provide better recommendation.

%By doing so, we demonstrate that auxiliary KCs can serve as a valuable intermediate representation, enabling improvements in multiple educational applications.% while maintaining model interpretability.

%We evaluate our model on real-world educational datasets to assess its effectiveness in improving predictive accuracy and recommendation quality. The results demonstrate that incorporating auxiliary KCs learned through our model generally enhances the performance of both knowledge tracing models and can be utilized to provide better recommendation algorithms.

The contributions of this paper are threefold:
\begin{itemize}
    \item We introduce a model that learns sparse binary representations for exercises, from which auxiliary KCs are derived.
    \item We show how these auxiliary KCs improve classical knowledge tracing such as BKT.
    \item We introduce two recommendation algorithms that can take advantage of these learned representation to provide better recommendation.
    \item We perform extensive experiments on multiple datasets, demonstrating the effectiveness of our approach across multiple tasks.
\end{itemize}

%In the following sections, we review related work in knowledge tracing and representation learning, describe our method for extracting and integrating auxiliary KCs, present experimental results, and discuss the implications for educational systems.

\section{Related Work}

%\subsection*{Knowledge Tracing Approaches}

One of the earliest approaches to knowledge tracing is Bayesian Knowledge Tracing (BKT) \cite{bkt}, which models student understanding of individual knowledge components (KCs) using a Hidden Markov Model (HMM). BKT offers high interpretability and simplicity. However, it assumes independence among KCs, which restricts its ability to capture complex interdependencies. Consequently, its predictive performance often falls short when compared to modern deep learning-based approaches \cite{whendeep,Howdeep}.

Deep Knowledge Tracing (DKT) \cite{dkt} introduced recurrent neural networks (RNNs) to model the temporal evolution of student knowledge. Following the introduction of DKT, a wave of subsequent models began leveraging deep learning architectures. Attention-based models such as SAKT \cite{sakt} and memory-augmented architectures like DKVMN \cite{dkvmn}.

The Dynamic Key-Value Memory Network (DKVMN) \cite{dkvmn} employs a memory-augmented framework comprising a static key memory and a dynamic value memory. The key memory captures latent concepts by learning fixed relationships between exercises and conceptual structures, while the value memory dynamically tracks a student's mastery over these concepts. However, these concepts are only used internally by the model and not as a learned representation to be used in other tasks.%Unlike traditional methods that rely on human-annotated KCs, DKVMN infers latent conceptual representations directly from the data, bypassing the need for predefined tags.

Several models in the literature diverge in their treatment of KCs. Some model operate without relying on human-defined KCs and instead use learned question embeddings, such as DKVMN. Others, such as QIKT\cite{qikt}, incorporate both question information and KCs to leverage the strengths of both representations.

Given that human-defined KCs can be noisy or incomplete, a number of studies focus on refining these annotations. Some methods attempt to calibrate or correct the original KC assignments by learning adjustments to the question-to-concept mappings \cite{calib_q_matrix,calib_q_matrix2,calib_q_matrix3}. These works aim to improve the quality of KC annotations, often treating expert-provided KCs as an initial approximation. Our work, however, takes a different approach. Rather than modifying existing KCs, we introduce auxiliary KCs that complement the original annotations. 

Other research efforts propose learning dense embeddings for use in downstream KT models \cite{pre_cont_embed2020,cont_emb2}. While effective in capturing semantic relationships, these dense embeddings often lack interpretability and are generally restricted to deep learning applications. In contrast, our method learns discrete representation which is inherently more interpretable and compatible with both classical and deep learning models.

The work in \cite{end2end_binary} proposes learning new KC representations to replace those defined by experts. Each question is represented by a binary vector, where each dimension denotes the presence or absence of a KC. However, the model does not directly produce binary representations. Instead, it learns dense vectors and applies regularization to approximate binary behavior, followed by post-hoc binarization. In our method, binary representations are learned explicitly during training, yielding discrete auxiliary KCs without the need for approximation. Although our framework can be extended to fully substitute the original KCs, our main objective is to augment the original KCs rather than replace them.

  %Recommendation approaches in education can include rule-based systems, collaborative filtering, and matrix factorization, which primarily rely on static user profiles or performance histories. While effective in some settings, these methods often fail to capture the sequential and dynamic nature of student learning.

%%%%%%%%

In adaptive learning, reinforcement learning (RL) has emerged as a promising framework for developing recommendation policies. A common approach is to model the recommendation process as a Partially Observable Markov Decision Process (POMDP) as the true knowledge state of the student is hidden while decisions are made based on observable interaction histories \cite{sergey,dkvmn-ca}. This formulation highlights the critical role of accurate student modeling in educational recommender systems.

In parallel, non-RL methods have also been explored. For example, the approach in \cite{huo2020knowledge}—termed “recommending towards weaknesses”—first identifies the weakest knowledge component (KC) that meets a minimum threshold of deficiency, and then recommends an exercise strongly associated with that KC. However, an expert is required to tune the threshold instead of learning it. %In \cite{dkt}, they propose an expectimax algorithm, which selects the KC that is expected to yield improvement according to the student model. This approach 

%Regardless of whether an RL-based or non-RL-based strategy is employed, an effective student model remains central to adaptive learning systems, as it informs decisions that guide learners toward optimal educational outcomes.

\section{BACKGROUND}

In this section, we provide an overview of knowledge tracing models that are relevant to this work. 
\subsection{Bayesian Knowledge Tracing}

Bayesian Knowledge Tracing (BKT)~\cite{bkt} is a Hidden Markov Model (HMM), where a learner's knowledge state of each KC is treated as a hidden (latent) variable. The framework is designed to estimate whether a student has mastered a particular KC by observing their responses to practice opportunities.

BKT models the learning process using an HMM as follows:
\begin{itemize}
    \item \textbf{Latent Knowledge State:} Indicates whether the learner has mastered the KC ($K_t = 1$) or has not ($K_t = 0$).
    \item \textbf{Observed Response:} Represents a correct ($O_t = 1$) or incorrect ($O_t = 0$) answer at time step $t$.
\end{itemize}

 The model is defined using the following key parameters:
\begin{itemize}
    \item $P(L_0)$: Initial probability that the learner has already mastered the KC prior to any practice.
    \item $P(T)$: Learning probability, which denotes the chance of transitioning from non-mastery ($K_t = 0$) to mastery ($K_{t+1} = 1$) after an opportunity to practice.
    \item $P(G)$: Guess probability, the likelihood of answering correctly despite not having mastered the KC.
    \item $P(S)$: Slip probability, the likelihood of an incorrect response despite having mastered the KC.
\end{itemize}

The primary transitions in the HMM are as follows:
\begin{itemize}
    \item Learners may transition from non-mastery to mastery with probability $P(T)$.
    \item Once mastery is achieved, it is considered \textbf{absorbing}, meaning the learner remains in the mastered state permanently.
\end{itemize}

The standard BKT framework assumes that KCs are independent; mastery of one KC does not affect mastery of another. While this assumption simplifies the model, it does not always reflect real-world learning contexts where skills are often interdependent. Additionally, BKT does not account for forgetting; once a KC is mastered, the probability of reverting to non-mastery is assumed to be zero. A more detailed explanation of the BKT model can be found in~\cite{bkt_intro,bkt}.

\subsubsection{BKT with Forgetting}

A widely used extension of BKT incorporates the concept of forgetting, allowing transitions from mastery back to non-mastery. Several approaches have been proposed to model this behavior~\cite{BKTvariants,pybkt}. One such method, as described in~\cite{Howdeep}, introduces a forgetting probability $P(F)$, defined as follows:

\begin{equation}
P(K_{t+1} = 0 \mid K_t = 1) = P(F)
\end{equation}

This variant captures the possibility that learned knowledge may decay over time. In this work, we use this forgetting-aware version of BKT exclusively. Unless otherwise noted, all references to BKT in the following sections refer to this specific variant.

\subsection{Deep Knowledge Tracing}
\label{sec:dkt}

Deep Knowledge Tracing (DKT) \cite{dkt} leverages a recurrent neural network (RNN) to model sequences of learner interactions and forecast future performance. Each input sequence comprises interaction pairs \((q_t, y_t)\), where \(q_t\) denotes a single KC in the original formulation, and \(y_t\) indicates the correctness of the learner’s response (\(y_t = 1\) if correct, \(y_t = 0\) if incorrect).

The RNN iteratively updates a hidden state \(h_t\) that captures the learner's evolving knowledge. This is defined as:
\begin{equation}
h_t = f(h_{t-1}, x_t),
\end{equation}
where \(x_t = (q_t, y_t)\) encodes the interaction at time \(t\), and \(f\) represents the RNN update function, such as an LSTM or GRU.

Given the updated hidden state \(h_t\), the model estimates the probability of a correct response for each KC at each time step, achieved by passing \(h_t\) through a fully connected output layer followed by a softmax activation function.

Although DKT typically outperforms traditional models like Bayesian Knowledge Tracing (BKT) \cite{Howdeep,whendeep} and relies on fewer modeling assumptions, it has several limitations. A major concern is its limited interpretability. Unlike BKT, which offers interpretable parameters such as learning and slip probabilities, the hidden state in DKT lacks transparency and does not readily yield actionable insights.

%Moreover, DKT is prone to label leakage, necessitating careful evaluation protocols to ensure valid performance estimates \cite{pykt,leak}. To avoid this issue, we adopt the variant proposed in \cite{leak}, where \(q_t\) denotes the set of knowledge components associated with a given question, and \(x_t\) is computed as the average of their corresponding embeddings. Throughout this work, all references to DKT pertain to this modified formulation.

Additionally, the original implementation of DKT is susceptible to label leakage, requiring careful evaluation strategies \cite{pykt,leak}. To mitigate this, we adopt a variant from \cite{leak} in which \(q_t\) represents the set of KCs associated with a given question, and \(x_t\) is defined as the mean of the embeddings of these KCs. Throughout this paper, we refer to this variant when mentioning DKT.

%DKT represents a significant advancement in the field of knowledge tracing, offering a flexible and expressive framework for modeling student learning. Nonetheless, challenges related to interpretability and data requirements continue to underscore the value of classical models such as BKT. This work aims to bridge these approaches by transferring the representational strengths of deep learning into interpretable frameworks like BKT.

\subsubsection{The Expectimax exercise recommendation algorithm}
\label{sec:expectimax}

In \cite{dkt}, they suggested the use of their DKT model to provide recommendation by doing a one step exploration, calling it an Expectimax algorithm. At each time step, all possible KCs are tested on the DKT. Since DKT outputs probabilities for all KCs at each time step, it's possible to measure which choice has the higher effect on improving the student results on all KCs. Thus, the KC that gives the highest improvement on this student model is recommended. Later, an exercise that is associated with the chosen KC can be recommended. Still, this leaves a large set of possible exercises to chose from. To mitigate this, we create a variant of this algorithm that utilize auxiliary KCs to narrow down the choices, in Section~\ref{sec:aux_expectimax}.

%However, directly testing all the exercises is unfeasible. Instead, the smaller set of KCs are explored instead of the exercises. The algorithm then recommend a KC instead of an exercise.% In this work, we will investigate if our learned auxiliary KCs

\section{Model Overview}

\begin{figure}[!ht]
\centering
\includegraphics[width=0.4\linewidth]{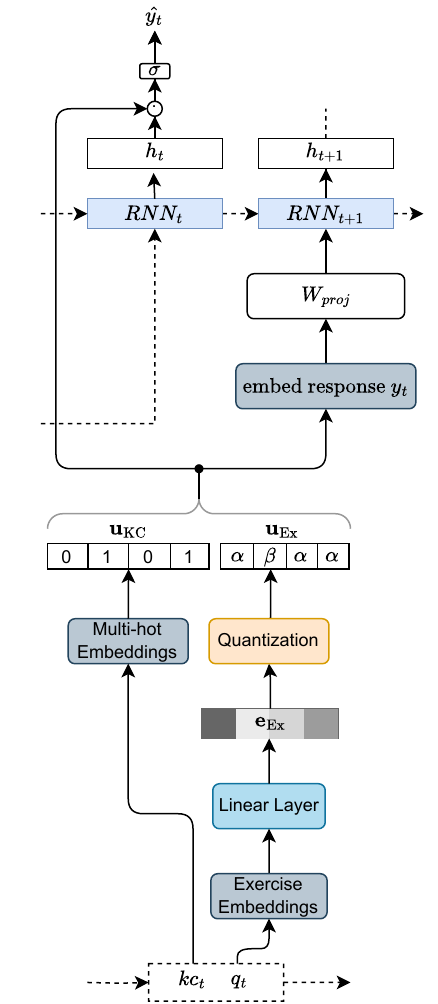}
\caption{Architecture of the proposed model. The figure is adapted from \cite{csedu}.}
\label{fig:main}
\end{figure}

SBRKT is a model that combines predefined and learnable auxiliary KCs to trace student knowledge over time. The core idea is to represent each exercise as a sparse binary vector, where each bit denotes the presence or absence of an auxiliary KC. These vectors are then used as inputs to an RNN for prediction.

\subsection{Multi-Hot Representation of Human-Defined KCs}

Let \( N \) denote the number of predefined KCs. For each exercise, we construct a binary vector \( \mathbf{u}_{\text{kc}} \in \{0,1\}^N \), where:
\[
\mathbf{u}_{\text{kc}}[i] = \begin{cases}
1, & \text{if the } i\text{-th KC is associated with the exercise}, \\
0, & \text{otherwise}.
\end{cases}
\]

To incorporate the correctness label \( y \in \{0,1\} \), we define a labeled vector \( \mathbf{u}_{\text{kc}, y} \in \{0,1\}^{2N} \) as:
\[
\mathbf{u}_{\text{kc}, y} = y \cdot \mathbf{u}_{\text{kc}} \oplus (1 - y) \cdot \mathbf{u}_{\text{kc}},
\]
where \( \oplus \) denotes vector concatenation. The first \( N \) components represent correct responses, and the last \( N \) components encode incorrect ones.

\subsection{Binary Encoding via Exercise Embeddings}

Each exercise \( q \) is embedded as an embedding vector \( \mathbf{x}_q \in \mathbb{R}^d \), which is passed through a linear layer:

\begin{equation}
\label{eq:linear}    
\mathbf{e}_q = \mathbf{W} \mathbf{x}_q + \mathbf{b}, \quad \mathbf{W} \in \mathbb{R}^{M \times d}, \quad \mathbf{b} \in \mathbb{R}^M,
\end{equation}
where \( M \) is the number of learned auxiliary KCs.

\subsubsection{Sparse Binary Quantization}

The goal is to discretize \( \mathbf{e}_q \in \mathbb{R}^M \) into a vector representation that can be later reduced into a binary vector. The output of our algorithm is \( \mathbf{u}_q \in \{\alpha, \beta\}^M \), where \( \alpha > \beta \). Let \(C_{\max}\) be a hyperparameter that represent the maximum number of auxiliary KCs per exercise. We output $u_q$ through the following steps: 
%and only the top \( C_{\max} \) elements are activated (non-zero).

\begin{enumerate}
    \item \textbf{Top-\( C_{\max} \) Selection:} Let \( \mathcal{I}_{\text{top}} \subseteq \{1,\dots,M\} \) be the indices of the top \( C_{\max} \) entries in \( \mathbf{e}_q \). Define a mask \( \mathbf{m} \in \{0,1\}^M \) as:
    \[
    \mathbf{m}[i] = \begin{cases}
    1, & \text{if } i \in \mathcal{I}_{\text{top}}, \\
    0, & \text{otherwise}.
    \end{cases}
    \]

    \item \textbf{Discretization:} Define a thresholding function \( f \colon \mathbb{R} \to \{0,1\} \) as:
    \[
    f(x) = \begin{cases}
    1, & \text{if } x > 0, \\
    0, & \text{otherwise}.
    \end{cases}
    \]
    Apply \( f \) elementwise and multiply by the mask:
    \[
    \mathbf{q} = f(\mathbf{e}_q) \odot \mathbf{m}.
    \]

    \item \textbf{Binary Mapping:} Map to scalar values \( \alpha = c(1 + \sigma(p_\alpha)) \), \( \beta = c \cdot \sigma(p_\beta) \), where $c$ is a hyperparameter that is set to one in our implementation. Both $p_\alpha$ and $p_\beta$ are scalar learnable parameters. Lastly, we generate a vector that has one-to-one mapping with a binary vector as follows:
    \[
    \mathbf{u}_q = \alpha \cdot \mathbf{q} + \beta \cdot (1 - \mathbf{q}).
    \]
\end{enumerate}

To train the discretization part of the  model, we use the Straight-Through-Estimator (STE)\cite{ste} which treats the quantization as identity in the backward pass.
%\[
%\frac{\partial Q(\mathbf{e}_q)}{\partial \mathbf{e}_q} = \mathbf{I}.
%\]

\subsubsection{Embedding Correctness Labels into Auxiliary KCs}

To encode correctness labels we use the same approach used for human-labeled KCs. We encode the binary exercise vector as:
\[
\mathbf{u}_{q, y} = y \cdot \mathbf{u}_q \oplus (1 - y) \cdot \mathbf{u}_q \in \mathbb{R}^{2M}.
\]

\subsection{Temporal Modeling}

At each time step \( t \), a feature vector is formed by concatenating the label-encoded KC embeddings and label-encoded quantized embedding as follows:
\[
\mathbf{v}_t = \mathbf{u}_{\text{kc}, y_t} \oplus \mathbf{u}_{q, y_t} \in \mathbb{R}^{2N + 2M}.
\]

We project \( \mathbf{v}_t \) into a dense vector:
\[
\mathbf{z}_t = \mathbf{W}_{\text{proj}} \mathbf{v}_t, \quad \mathbf{W}_{\text{proj}} \in \mathbb{R}^{D \times (2N + 2M)}.
\]

An RNN processes the sequence \( \{\mathbf{z}_1, \dots, \mathbf{z}_T\} \), updating the hidden state:
\[
\mathbf{h}_t = \mathrm{RNN}(\mathbf{z}_t, \mathbf{h}_{t-1}).
\]

Finally, logits are computed for all KCs as follows:
\[
\mathbf{o}_t = \mathbf{W}_{\text{out}} \mathbf{h}_t + \mathbf{b}_{\text{out}}, \quad \mathbf{W}_{\text{out}} \in \mathbb{R}^{(N+M) \times D}.
\]

To predict the response, we concatenate the binary KC and exercise vectors to perform a dot product with the output of the RNN:
\[
\mathbf{u}_t = \mathbf{u}_{\text{kc},t} \oplus \mathbf{u}_{q,t} \in \{0,1\}^{N+M},
\]
\[
\hat{y}_t = \sigma(\mathbf{u}_t^\top \mathbf{o}_t),
\]
where \( \sigma(\cdot) \) is the sigmoid function which is applied to output probabilities.

\subsection{Using Binary Representations in Downstream Models}

The quantized vector \( \mathbf{u}_q \in \{\alpha, \beta\}^M \) contains only two values, $\alpha$ and $\beta$ such that \( \alpha > \beta \). We map \( \alpha \to 1 \) and \( \beta \to 0 \) to produce a standard binary encoding of auxiliary KCs, which can be directly utilized in models such as BKT and DKT.

\section{Exercise Recommendation Algorithms}

In this section, we explore how the learned auxiliary KCs can be integrated into KT-based recommendation algorithms to improve exercise selection.

\subsection{Expectimax with Auxiliary KCs}
\label{sec:aux_expectimax}

To incorporate auxiliary KCs into the Expectimax algorithm, we extend the recommendation process to consider both human-defined and learned auxiliary KCs. Specifically, we perform Expectimax planning independently over each KC set (human defined KCs and learned auxiliary KCs), then intersect the exercises associated with each KC. That is, we select exercises that are associated with both a human-defined KC and an auxiliary KC.

This dual-filtering mechanism helps to narrow the set of candidate exercises, allowing for more targeted and precise recommendations. By leveraging the complementary strengths of human annotations and data-driven representations, this approach enhances the granularity of the recommendation process.

\subsection{Reinforcement Learning Based Exercise Recommendation}
To leverage the extracted auxiliary KCs, we design a reinforcement learning policy based on Deep Knowledge Tracing (DKT). Specifically, we adopt the Proximal Policy Optimization (PPO) algorithm. The policy network architecture is similar to DKT except for the output layer, which is replaced by the output of all possible actions (the exercises). Adapting DKT helps solve the partially observe nature of learning, where the actual student knowledge state is not observed \cite{sergey,dkvmn-ca}. 

For the reward design, we use the approach in \cite{dkvmn-ca} where the reward is the percentage of questions in the dataset that a student can answer at each time step.% Thus, the policy aims to learn the long-term reward of 

%The agent operates within a simulated student learning environment and aims to optimize long-term learning outcomes, as measured by the percentage of correctly answered questions across the entire dataset. By observing students’ historical interactions, the agent learns to recommend personalized sequences of exercises that maximize overall learning effectiveness.

\subsubsection{Model Architecture}

Our model incorporates KCs associated with each exercise. The architecture consists of the following key components:

\begin{itemize}
    \item \textbf{Input Representation:} At each timestep, the agent receives a pair $(q_t, r_t)$, where $q_t$ is the exercise attempted and $r_t \in \{0, 1\}$ is the correctness of the student's response. The exercise is mapped to its associated KCs.
    
    \item \textbf{Knowledge Tracing (DKT):} We employ similar architecture to DKT as described in Section~\ref{sec:dkt}. The only difference is the lack of the output layer. We only use the hidden state $h_t$ of the LSTM. 

    \item \textbf{Policy and Value Heads:} The output of the LSTM is fed into two separate fully connected layers: one for the policy network (actor) and one for the value network (critic). The actor produces logits over the action space (i.e., the set of available exercises), from which the next recommended exercise is sampled. The critic estimates the value of the current state for PPO optimization.

\end{itemize}

%\subsection{Training with PPO}

%The agent is trained using the PPO algorithm, which balances exploration and exploitation by constraining policy updates. At each training step, the agent samples trajectories by interacting with the environment, receives rewards based on student performance, and updates its parameters to maximize expected cumulative reward. The reward is defined as the percentage of correct responses over the episode, providing a direct signal aligned with learning objectives.

\subsection{Evaluation: Student Simulation Environment}

To simulate student learning behavior and enable interaction with the reinforcement learning agent, we employ the DKVMN model. DKVMN demonstrated strong performance in modeling student knowledge by explicitly modeling student memory while being KC agnostic as the agent should recommend exercises in production not KCs.

%We let the agent assess the utility of the extracted auxiliary KCs

To enhance the simulation accuracy, we incorporate prediction-consistent regularization techniques from DKT+ \cite{dkt_plus}. This helps prevent the model from showing unrealistic output such as fluctuation across time steps that is inconsistent with the student historical performance. Specifically, they add regularization terms that encourage the model to stay close to recent interactions and produce smoother transitions in predicted mastery levels. This results in more realistic student behavior, providing a stable and interpretable environment for training the recommendation policy.

\section{EXPERIMENTS}
In this section, we evaluate the effectiveness of SBRKT model and the utality of its learned representation. First, we compare SBRKT performance with some baseline models. Later, we use its extracted auxiliary KCs in downstream tasks. The first downstream task is basic student modeling, in which we use the unaltered BKT model on the same data but with the added auxiliary KCs. The second downstream task is exercise recommendation, where we apply the auxiliary KCs to both a reinforcement learning algorithm and to an Expectimax algorithm.

\subsection{Experimental Setup}

\subsubsection{Datasets}

We select publicly available datasets commonly used in knowledge tracing research:
\begin{itemize}
    \item \textbf{ASSISTments2009}\footnote{\url{https://sites.google.com/site/assistmentsdata/home/}}: This dataset originates from the ASSISTments online learning platform and was collected during the 2009–2010 academic year. Of the two versions provided, we use the skill-builder dataset.
    
    \item \textbf{ASSISTments2017}\footnote{\url{https://sites.google.com/view/assistmentsdatamining/}}: A more recent dataset from ASSISTments, released for the Workshop on Scientific Findings from the ASSISTments Longitudinal Data Competition at the 11th Conference on Educational Data Mining. We utilize the publicly available preprocessed version from \cite{akt}.
    
    \item \textbf{Algebra2005} \cite{algebra05}: This dataset was featured in the 2010 KDD Cup, part of the Educational Data Mining Challenge.
    
    \item \textbf{riiid2020} \cite{riiid}: Released as part of a Kaggle competition focused on enhancing AI-based student performance prediction. It contains millions of anonymized student interactions with an AI tutoring system centered on question-solving. We sample one million interactions from this dataset for our use.
\end{itemize}

Detailed statistics about the datasets can be found in table~\ref{table:stats}

\begin{table}[h]
\caption{Datasets features after prepossessing.}%\vspace*{1ex}
\label{table:stats}
\centering
\begin{tabular}{llllll}
\toprule% \\
\textbf{dataset} & questions & KCs  & students \\ \midrule
ASSISTments2009         & 17751           & 123           & 4163               \\
ASSISTments2017         & 3162           & 102           & 1709               \\
Riiid2020       & 13522          & 188          & 3822                        \\ 
Algebra2005        & 173650          & 112           & 574                                  \\
\end{tabular}
\end{table}

\subsubsection{Baselines}

We evaluate our model on the KT task against the following baseline methods:
\begin{itemize}
    \item \textbf{Bayesian Knowledge Tracing (BKT)}\cite{bkt}.
    
    \item \textbf{Deep Knowledge Tracing (DKT)}.
    
    \item \textbf{Dynamic Key-Value Memory Networks (DKVMN)}\cite{dkvmn}: A memory-augmented neural model that uses two memory types—key memory to represent latent knowledge concepts and value memory to track student performance.
    
    \item \textbf{Deep Item Response Theory (deepIRT)}\cite{deepirt}: An extension of DKVMN that integrates Item Response Theory (IRT), a psychometric framework that captures the interplay between student ability, question difficulty, and the likelihood of a correct response.
    
    \item \textbf{Question-centric Interpretable Knowledge Tracing (QIKT)}\cite{qikt}: A deep learning model that also incorporates IRT, focusing on providing interpretability by modeling question-centric features.
\end{itemize}

\subsubsection{Implementation Details}

For our proposed model, we set the embedding dimension to $d = 32$ for both the dense and binary exercise embeddings, corresponding to a max of $32$ auxiliary knowledge components (KCs). We adopt an LSTM architecture for the recurrent neural network, with a hidden state size of $h = 128$. Each exercise is associated with up to $C_{\text{max}} = 4$ auxiliary KCs.

All deep-learning KT models are trained using the Adam optimizer with a learning rate of $0.001$.  BKT is trained using stochastic gradient descent with a learning rate of $0.01$. The batch size is set to $32$ for DKVMN, deepIRT, and QIKT, and $128$ for DKT and BKT.

Experiments are conducted using an 80/10/10 split for training, validation, and testing, respectively. Model performance is evaluated using the Area Under the Curve (AUC) metric.

\subsection{Results}

\subsubsection{Performance Comparison}

Table~\ref{table:pre} presents a comparison of our proposed model against the baseline methods. The results show that our model achieves the highest performance on several datasets and ranks second on the others. These outcomes highlight the effectiveness of our approach on the KT task alone. Even under discrete constraints, our model can outperform alternatives that rely on dense representations.

%For a fair comparison, we fixed the maximum number of auxiliary KCs for all experiments, $C_{\text{max}} = 4$. However, in practice, hyperparameters would be tuned differently for each dataset.

\begin{table*}[!ht]
\centering
\caption{AUC Scores with top performers highlighted (* Best, ** Second Best) \cite{csedu}.}
\label{table:pre}
\begin{tabular}{lllll}
\toprule
Model & Algebra2005 & ASSISTment2009 & ASSISTment2017 & riiid2020 \\
%Model &  &  &  &  \\
\midrule
BKT & 0.7634 & 0.6923 & 0.6081 & 0.6215 \\
DKT & 0.8198 & 0.7099 & 0.6807 & 0.6503 \\
DKVMN & 0.7759 & 0.7362 & 0.7169 & \textit{0.7362}** \\
DeepIRT & 0.7750 & 0.7374 & 0.7170 & 0.7360 \\
SBRKT & \textit{0.8223}** & \textbf{0.7602}* & \textit{0.7494}** & \textbf{0.7369}* \\
QIKT & \textbf{0.8335}* & \textit{0.7574}** & \textbf{0.7527}* & 0.7324 \\
\bottomrule
\end{tabular}
\end{table*}

\subsubsection{Downstream Task Performance with BKT}

\begin{table*}[bp]
\centering
\caption{AUC Scores with Top Models Highlighted (* Best, ** Second Best). DKT+aux and BKT+aux indicate DKT and BKT models augmented with pretrained auxiliary KCs \cite{csedu}.}
\label{table:downstream}
\centering
\begin{tabular}{lllll}
\toprule
Model & Algebra2005 & ASSISTment2009 & ASSISTment2017 & riiid2020 \\
\midrule
BKT & 0.7634 & 0.6923 & 0.6081 & 0.6215 \\
BKT+aux & 0.7655 & \textit{0.7325}** & 0.6760 & \textit{0.7173}** \\
DKT & \textbf{0.8198}* & 0.7099 & \textit{0.6807}** & 0.6503 \\
DKT+aux & \textit{0.7997}** & \textbf{0.7481}* & \textbf{0.7422}* & \textbf{0.7365}* \\
\bottomrule
\end{tabular}
\end{table*}

We evaluate the effectiveness of the extracted auxiliary KCs by using them as input features for both BKT and DKT models, as summarized in Table~\ref{table:downstream}. The results indicate that BKT enhanced with auxiliary KCs (BKT+aux) outperforms the standard DKT on the ASSISTments2009 and Riiid2020 datasets. Additionally, incorporating auxiliary KCs consistently improves BKT's performance across all datasets, though the gain on Algebra2005 is marginal. On the other hand, DKT augmented with auxiliary KCs (DKT+aux) shows improved performance on all datasets except Algebra2005, where it performs slightly worse than the original DKT.

%latent KCs significantly improve the performance of BKT, demonstrating their utility as disentangled and meaningful representations.
\subsection{Recommender Algorithms}

To evaluate the the introduced recommender algorithms, we use 24 simulated students. Each student will work with 140 exercise recommended by the algorithm. We use two evaluation metrics. The first is the average reward over the fixed-length experiment (140 exercise), which is the total percentage of exercises that a student was able to answer over the whole plan.

The second metric is based on the average normalized gain \cite{gain}, defined as the ratio of the actual gain to the maximum possible gain. For a given pre-test score \( s_{\text{pre}} \) and post-test score \( s_{\text{post}} \), the normalized gain \( G_1 \) is calculated as:

\begin{equation}
G_1 = \frac{s_{\text{post}} - s_{\text{pre}}}{1 - s_{\text{pre}}}
\end{equation}

\noindent where:
\begin{itemize}
  \item \( s_{\text{pre}} \) is the students’ average score before applying the algorithm plan,
  \item \( s_{\text{post}} \) is the average score at the end of the algorithm plan,
  \item \( 1 \) represents the maximum possible score (i.e., all questions answered correctly).
\end{itemize}

Our PPO implementation is adapted  from CleanRL \cite{cleanrl}. We use a learning rate of $2.5 \times 10^{-4}$.

To evaluate the re commender algorithms, we train our simulator on the \textit{ASSISTments2009} dataset. For each algorithm, we apply it to 24 random students. Our experiments shows consistent improvement of adding the pre-trained auxiliary KCs to the PPO based policy as seen in Table~\ref{tab:rl}. Moreover, incorporating auxiliary KCs into the Expectimax algorithm can show improvements as seen in Table~\ref{tab:rl}. %However, we note that this improvement is less consistent than with the RL based algorithms.

\begin{table}[ht]
\centering
\caption{Performance comparison of PPO-LSTM and Expectimax on the AS09 dataset with and without auxiliary KCs (aux-KCs). Aux-KCs refer to learned representations obtained via pre-training.}
\label{tab:rl}
\begin{tabular}{lccc}
\hline
\textbf{Method} & \textbf{Aux-KCs} & \textbf{Gain} & \textbf{Mean} \\
\hline
\multirow{2}{*}{PPO-LSTM}
& Yes & \textbf{0.6331} & $\mathbf{0.8613 \pm 0.0584}$ \\
& No & 0.2417 & $0.6963 \pm 0.0759$ \\
\hline
\multirow{2}{*}{Expectimax}
& Yes & 0.4227 & $0.7484 \pm 0.0802$ \\
& No & 0.2171 & $0.6805 \pm 0.0745$ \\
\hline
\end{tabular}
\end{table}

\subsection{Ablation Study}

To evaluate the contribution of the quantization layer, we design three model variants:

\begin{itemize}
    \item \textbf{SBRKTtanh}: This variant introduces a hyperbolic tangent (tanh) activation function following the linear transformation in equation~\ref{eq:linear}. The output is discretized to either $-1$ or $+1$, rather than the standard $\alpha$ and $\beta$ values.
    
    \item \textbf{SBRKT10}: This variant applies a sigmoid activation to the linear transformation output. Values below 0.5 are mapped to 0, and values equal to or above 0.5 are mapped to 1, again replacing $\alpha$ and $\beta$.
    
    \item \textbf{SBRKTdense}: This version removes the quantization layer entirely and uses a continuous dense representation. However, such representations are incompatible with the downstream procedure outlined in this work.
\end{itemize}

As shown in Table~\ref{table:pre_ablation}, our proposed model achieves superior performance across most datasets. The only exception is the \textit{algebra2005} dataset, where it ranks second with a negligible margin (AUC difference of 0.006). Notably, the \textit{SBRKTdense} variant underperforms significantly, underscoring the importance of the quantization layer in the model’s effectiveness.

\begin{table}[H]
\centering
\caption{AUC Scores with Top Models Highlighted (* Best, ** Second Best). algebra05, assist09, and assist17 refer to Algebra2005, ASSISTments2009, and Assistments2017, respectively. SBR, SBR10, and SBRtanh denote SBRKT, SBRKT10, and SBRKTtanh, respectively \cite{csedu}.}
\label{table:pre_ablation}
\begin{tabular}{llll}
\toprule
Dataset & algebra2005 & assist09 & assist17 \\
\midrule
SBR & \textit{0.8223}** & \textbf{0.7602}* & \textbf{0.7494}* \\
SBR10 & \textbf{0.8231}* & \textit{0.7464}** & 0.7431 \\
SBRdense & 0.8122 & 0.7169 & 0.7448 \\
SBRtanh & 0.8166 & 0.7449 & \textit{0.7491}** \\
\bottomrule
\end{tabular}
\end{table}

To further assess the impact of these variants, we evaluate their extracted auxiliary KCs in enhancing other KT models. As reported in Table~\ref{table:downstream_ablation}, the auxiliary KCs extracted by our SBRKT model consistently lead to better performance across datasets. The only exception is the \textit{algebra2005} dataset, where no variant demonstrates a significant improvement, and in some cases, performance even declined with the addition of auxiliary KCs.

\begin{table}[H]
\centering
\caption{AUC Scores with Top Models Highlighted (* Best, ** Second Best). algebra05, assist09, and assist17 refer to Algebra2005, ASSISTments2009, and Assistments2017, respectively. +AX10 and +AXtanh indicate training with auxiliary KCs from SBRKT10 and SBRKTtanh \cite{csedu}.}
\label{table:downstream_ablation}
\begin{tabular}{llll}
\toprule
Dataset & algebra05 & assist09 & assist17 \\
\midrule
BKT & 0.7634 & 0.6923 & 0.6081 \\
DKT & \textbf{0.8198}* & 0.7099 & 0.6807 \\
BKT+aux & 0.7655 & \textit{0.7325}** & 0.6760 \\
DKT+aux & \textit{0.7997}** & \textbf{0.7481}* & \textbf{0.7422}* \\
BKT+AX10 & 0.7745 & 0.7283 & 0.6577 \\
DKT+AX10 & 0.7899 & 0.7318 & \textit{0.7301}** \\
BKT+AXtanh & 0.6860 & 0.6736 & 0.5969 \\
DKT+AXtanh & 0.7454 & 0.6974 & 0.6722 \\
\bottomrule
\end{tabular}
\end{table}

\subsection{Summary of Findings}

Our experimental results yield the following key insights:

\begin{itemize}
    \item The proposed model outperforms all baselines on multiple benchmarks, despite operating under sparse, discrete constraints. These constraints enable the effective extraction of auxiliary knowledge components.
    
    \item The learned auxiliary KCs prove valuable for downstream tasks. In particular, BKT consistently benefits from their inclusion across all datasets. While DKT achieved substantial gains on some datasets (e.g., a 6\% AUC increase on \textit{ASSISTments2017}), but experienced a 
    drop in performance on \textit{algebra2005}.

    \item The learned auxilliary KCs helped perform a better Expectimax recommendation and provided a noticeable boost in performance for the PPO based algorithm. 
\end{itemize}

\section{Conclusion}

In this work, we introduced a novel method for augmenting student modeling and recommendation systems through the use of learned auxiliary knowledge components (auxiliary KCs). By leveraging deep learning techniques to generate sparse binary representations of exercises, we created interpretable and discrete features that complement human-defined KCs. These auxiliary KCs not only enhance classical models like BKT but also integrate seamlessly into modern architectures such as DKT and reinforcement learning-based recommendation systems.

Our experiments across multiple real-world educational datasets demonstrate the consistent effectiveness of incorporating auxiliary KCs in both predictive and adaptive tasks. Moreover, the learned representations yield improved performance in reinforcement learning-driven recommendation scenarios, highlighting their value in personalized learning applications.

Ultimately, our approach bridges the gap between interpretable, discrete modeling and the representational power of deep learning. It offers a scalable and architecture-agnostic solution that improves model accuracy, enhances recommendation quality, and maintains interpretability. 

\section{Conclusion}

In this work, we introduced a novel method for augmenting student modeling and recommendation systems through the use of learned auxiliary knowledge components (auxiliary KCs). By leveraging deep learning techniques to generate sparse binary representations of exercises, we created interpretable and discrete features that complement human-defined KCs. These auxiliary KCs not only enhance classical models like BKT but also integrate seamlessly into modern architectures such as DKT and reinforcement learning-based recommendation systems.

Our experiments across multiple real-world educational datasets demonstrate the consistent effectiveness of incorporating auxiliary KCs in both predictive and adaptive tasks. Moreover, the learned representations yield improved performance in reinforcement learning-driven recommendation scenarios, highlighting their value in personalized learning applications.

Ultimately, our approach bridges the gap between interpretable, discrete modeling and the representational power of deep learning. It offers a scalable and architecture-agnostic solution that improves model accuracy, enhances recommendation quality, and maintains interpretability.

\begin{credits}
\subsubsection{\ackname} We would like to thank the anonymous reviewers for their valuable comments and constructive feedback. This work was funded by the federal state of Baden-Württemberg as part of the Doctoral Certificate Programme "Wissensmedien" (grant number BW6{\_}10).

\subsubsection{\discintname}
The authors have no competing interests to declare that are relevant to the content of this article. 
\end{credits}
\bibliographystyle{splncs04}
\bibliography{mybibliography}
\end{document}